\documentclass[sigconf]{acmart}
\usepackage{subcaption}
\AtBeginDocument{%
  \providecommand\BibTeX{{%
    \normalfont B\kern-0.5em{\scshape i\kern-0.25em b}\kern-0.8em\TeX}}}

\copyrightyear{2020}
\acmYear{2020}

\setcopyright{rightsretained}
\begin{document}
\fancyhead{}

\title{Learning to Create Better Ads: Generation and Ranking Approaches for Ad Creative Refinement}

\author{Shaunak Mishra}
\affiliation{%
  \institution{Yahoo Research}
}
\email{shaunakm@verizonmedia.com}

\author{Manisha Verma}
\affiliation{%
  \institution{Yahoo Research}
}
\email{manishav@verizonmedia.com}

\author{Yichao Zhou}
\affiliation{%
  \institution{University of California, Los Angeles
  }
}
\email{yz@cs.ucla.edu}

\author{Kapil Thadani}
\affiliation{%
  \institution{Yahoo Research}
}
\email{thadani@verizonmedia.com}

\author{Wei Wang}
\affiliation{%
  \institution{University of California, Los Angeles}
}
  \email{weiwang@cs.ucla.edu}


\begin{abstract}
In the online advertising industry, the process of designing an ad creative (\emph{i.e.}, ad text and image) requires manual labor. Typically, each advertiser launches multiple creatives via online A/B tests to infer effective creatives for the target audience, that are then refined further in an iterative fashion. 
Due to the manual nature of this process, it is time-consuming to learn, refine, and deploy the modified creatives. Since major ad platforms typically run A/B tests for multiple advertisers in parallel,
we explore the possibility of collaboratively learning ad creative refinement via A/B tests of multiple advertisers. In particular, given an input ad creative, we study approaches to refine the given ad text and image by:
(i) generating new ad text, (ii) recommending keyphrases for new ad text, and (iii) recommending image tags (objects in image) to select new ad image.
Based on A/B tests conducted by multiple advertisers, we form pairwise examples of inferior and superior ad creatives, and use such pairs to train models for the above tasks. For generating new ad text, we demonstrate the efficacy of an encoder-decoder architecture with copy mechanism, which allows some words from the (inferior) input text to be copied to the output while incorporating new words associated with higher click-through-rate. For the keyphrase and image tag recommendation task, we demonstrate the efficacy of a deep relevance matching model, as well as the relative robustness of ranking approaches compared to ad text generation in cold-start scenarios with unseen advertisers.
We also share broadly applicable insights from our experiments using data from the Yahoo Gemini ad platform.
\end{abstract}


\copyrightyear{2020}
\acmYear{2020}
\acmConference[CIKM '20]{Proceedings of the 29th ACM International Conference on Information and Knowledge Management}{October 19--23, 2020}{Virtual Event, Ireland}
\acmBooktitle{Proceedings of the 29th ACM International Conference on Information and Knowledge Management (CIKM '20), October 19--23, 2020, Virtual Event, Ireland}\acmDOI{10.1145/3340531.3412720}
\acmISBN{978-1-4503-6859-9/20/10}


\keywords{Online advertising; A/B testing; sequence2sequence; ad creatives.\vspace{-5pt}}

\maketitle

\section{Introduction} \label{sec:introduction}
The image and text used for an online ad (collectively called an ad creative) can be influential in targeting online users on a large scale.
\begin{figure}[]
\centering
  \includegraphics[width=1 \columnwidth]{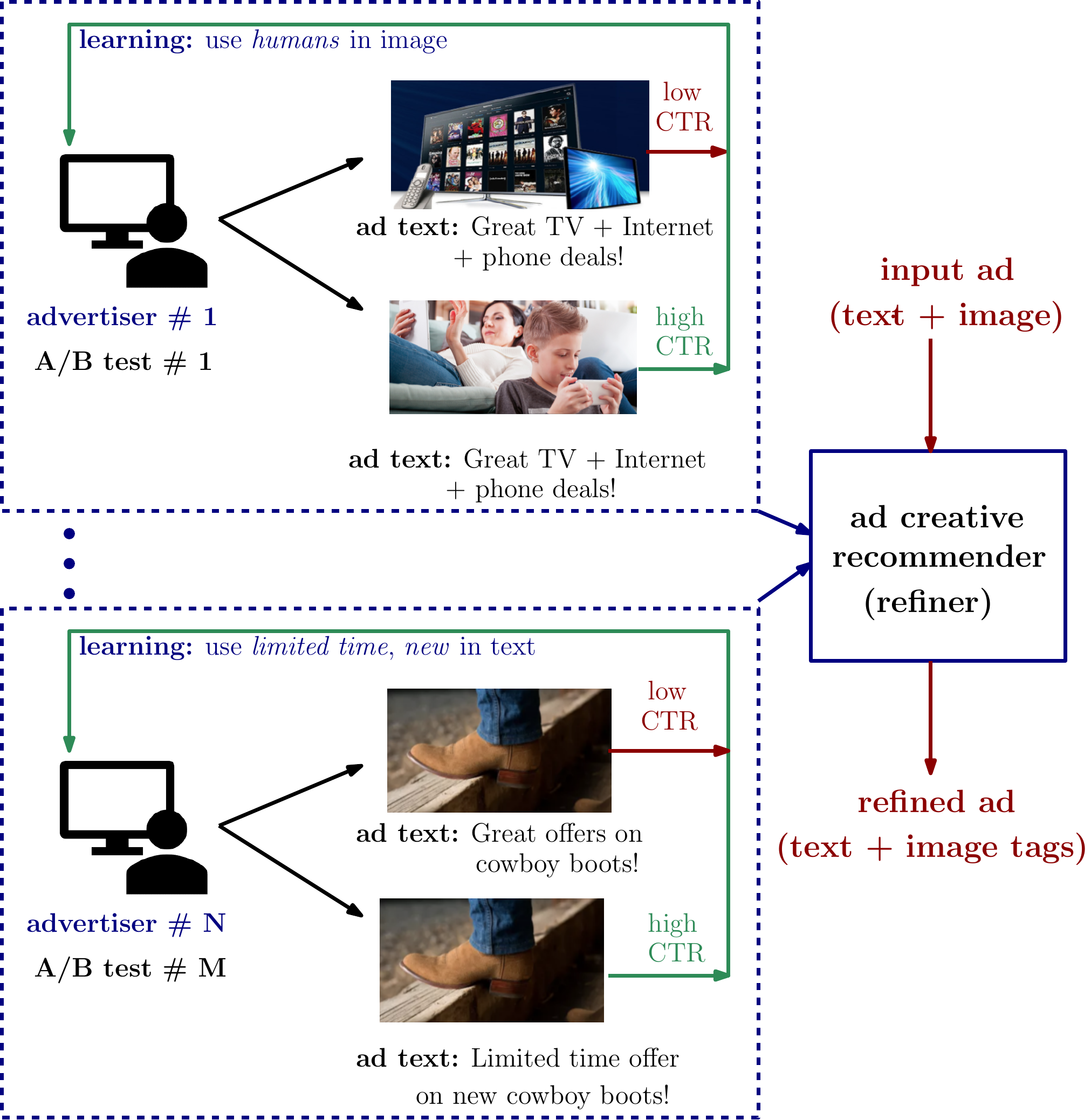}
  \caption{Ad creative refiner based on parallel A/B tests done by multiple advertisers. Advertiser $1$ may learn in isolation that having human elements leads to better CTR than multimedia images; while advertiser $N$ may learn that using "limited time" in ad text works better than "great". The proposed refiner collects data across A/B tests to recommend ad text and image refinements for a given input ad creative.}
  \label{fig:pull_figure}
\end{figure}
Large businesses (advertisers) typically employ creative strategists to design ad creatives; these creative strategists may conduct market research to see trending themes and also gather insights from past ad campaigns in related product categories.
Such advertiser specific creative customization is mostly a manual, expensive, and time consuming process. In contrast, small businesses typically resort to free online tools, \emph{e.g.}, stock image libraries \cite{shutterstock}, and generic creative insights \cite{taboola_trends} to compile ad images and text; such tools can reduce the time to design creatives but tend to be generic (\emph{e.g.}, lacking in business-specific customization). Once the ad creatives are ready, both large and small advertisers need to conduct online A/B tests to validate the effectiveness of their creatives, and subsequently discard low performing creatives from their ad campaigns. In addition, to reduce the chances of online users getting tired of seeing the same ad repeatedly on a particular website (\emph{i.e.}, ad fatigue \cite{ad_fatigue_schmidt}), advertisers need to frequently go through the design$\rightarrow$A/B test$\rightarrow$refresh ad creatives cycle 
. Again, such cycles tend to be time consuming and there is an emerging need for data-driven approaches to speed up the whole process of designing and refreshing creatives.
\par
In this paper, we highlight a key observation that accelerates the above creative design process, and can be explained as follows. Advertisers typically test their creatives via A/B tests in ad platforms (\emph{e.g.}, Yahoo Gemini, Facebook Ads), \emph{i.e.}, they try out a set of creatives on online users in a controlled setup such that the click-through-rate (CTR) performance \cite{mappi_CIKM} difference across the creatives can be solely attributed to the ad text and image. However, advertisers conduct and learn from such A/B tests in isolation as illustrated in Figure~\ref{fig:pull_figure}. As shown, advertiser $1$ who is an internet service provider, may learn via an A/B test that having human elements in the ad image works better than having gadgets in the image (since the ad text is same across the two creatives in the example, the performance difference can be attributed to the ad images). Via a separate A/B test, a different advertiser $N$ (selling boots) may learn that using \textit{new} and \textit{limited time} in the ad text works better than using \textit{great}. Our key observation in the illustrated example is that although the advertisers are learning in isolation, the ad platform can learn across advertisers. In fact, most ad platforms are authorized to use performance data across advertisers in an \textit{aggregate manner} to help advertisers perform better; however, using A/B test data across advertisers in a collaborative manner to automate ad creative refinement is a largely unexplored topic.

In this paper we address several sub-problems in ad creative (text and image) refinement exploiting the above observation by using multi-advertiser A/B test data:
\begin{enumerate}
\item ad text generation: given an input ad creative, the task is to generate refined ad text,
\item ad text keyphrase recommendation:
given an input ad creative, the task is to recommend keyphrases for inclusion in the refined ad text, and
\item ad image tag recommendation: given an input ad creative, the task is to recommend image tags (objects) to guide the selection of a refined ad image.
\end{enumerate}
Another novelty in our proposed approaches for the above tasks is that they do not depend on intermediate models such as CTR prediction as required in previous work \cite{microsoft_ad_generation_kdd19,mappi_CIKM} but rely on pairs of examples of the form: (low CTR creative, high CTR creative) where the CTR is based on the same population of users (\emph{i.e.}, targeting is fixed). Both creatives in a pair are sourced from the same advertiser, and at a high level, the task of refining can be seen as \textit{translating} the low CTR creative (source) to the high CTR creative (target). As we discuss in this paper, such pairs can be naturally collected from A/B tests conducted by multiple advertisers in an ad platform. 
Our main contributions are as follows.
\begin{itemize}
    \item We solve three tasks around ad creative refinement: (i) ad text generation, (ii) keyphrase recommendation, and (iii) image tag recommendation.
    \item For ad text generation, we demonstrate that using a copy mechanism to selectively copy parts of the input ad text while introducing new words in the refined (generated) text is significantly better than baselines.
  \item For keyphrase and image tag recommendation, we demonstrate the efficacy of a deep relevance matching model for ranking keyphrases and image tags. We also show the relative robustness of keyphrase ranking (compared to text generation) in a cold-start scenario with unseen advertisers. We observed a $87\%$ CTR increase via such recommendations for a major advertiser on Yahoo Gemini.
\end{itemize}
The remainder of the paper is organized as follows. Section~\ref{sec:related} covers related work, and Section~\ref{sec:problem_formulation} covers problem formulation. Section~\ref{sec:data} explains data sources, and creation of pairs of creatives for training ad refinement models.
Section~\ref{sec:method} covers proposed methods, Section~\ref{sec:results} covers experimental results, and there is a discussion in Section~\ref{sec:discussion}.

\section{Related Work} \label{sec:related}
\subsection{Online advertising}
Today, advertisers work with ad platforms \cite{mappi_CIKM,gemx_kdd,Google_FTRL} to launch campaigns that show ads to users on different websites. Advertisers design \emph{one or more creatives} with the help of creative strategists to target relevant online users and measure the effectiveness of campaigns with metrics such as click-through-rate (CTR $= \frac{clicks}{impressions}$) are associated with the ad creative being tested. It is common for advertisers to do exploratory A/B tests with a large pool of creatives to efficiently learn which creative works best (popularly known as dynamic creative optimization in the industry) \cite{explore_exploit_li}.
However, automatically understanding ad creatives (multi-modal in nature due to the presence of text and an image) and leveraging this understanding to create a pool of relevant creatives for A/B testing is emerging as an active area of research as described below.
Understanding content in ad images and videos from a computer vision perspective was first studied in \cite{cvpr_kovashka}, where manual annotations were gathered from crowdsourced workers for: ad category, reasons to buy products advertised in the ad, and expected user response given the ad.
Leveraging the dataset in this work, \cite{self_recsys2019} studied recommending keywords for guiding a brand's creative design. However, \cite{self_recsys2019} was limited to only text inputs for a brand (\emph{e.g.}, the brand's Wikipedia page), and the recommendation was limited to single words (keywords). In \cite{www20_joey}, the setup in \cite{self_recsys2019} was extended by including multi-modal information from past ad campaigns, \emph{e.g.}, images, text in the image (OCR), and Wikipedia pages of associated brands.
In this paper, we focus on refining existing (input) ad creatives, \emph{i.e.}, the refinement is specific to the input ad creative as opposed to providing recommendations for an input advertiser in \cite{self_recsys2019,www20_joey}. In addition, the usage of A/B test data across advertisers is another key difference with respect to prior work.
Our approaches are limited to consuming only CTR data across advertisers (and not conversions),
since in most cases, it is \textit{owned} by the ad platform, which is typically authorized to use aggregated data across advertisers to make system-wide improvements (not biased towards a particular advertiser).

\subsection{Relevance matching}
One of our goals is to recommend a set of highly relevant keyphrases and image tags for improving an (input) ad creative. This can be modeled as a query-document relevance ranking problem \cite{drmm}, or as a collaborative filtering problem where \textit{user-item} latent representations are used for recommendations \cite{koren_MF,neural_collaborative_filtering}.
However, given the restriction on the number of keyphrases/image tags recommended, and their relevance to the advertiser under consideration, we focus on relevance ranking models (\emph{e.g.}, DRMM \cite{drmm} and variants \cite{drmm_topk}) in our keyphrase and image tag ranking setups.

\subsection{Text-to-text generation}
We formulate ad text generation as a sequence-to-sequence prediction task, which is common in natural language processing problems like machine translation and abstractive summarization. State-of-the-art performance in machine translation is typically obtained with 
an encoder-decoder neural architecture with attention \cite{luong2015attention}. In abstractive summarization, where both the source and target sequences are in the same language, an additional mechanism to copy input tokens to the output sequence has proven to be beneficial \cite{see_pointer_generator}. In the context of ad text generation, recent work \cite{microsoft_ad_generation_kdd19} explored the use of an encoder-decoder architecture to automatically generate ad text based on an advertiser's webpage. The main differences between our work and \cite{microsoft_ad_generation_kdd19} lie in: (i) studying ad refinement as opposed to generating an ad from scratch, (ii) the use of A/B test data across advertisers to train refinement models.

\section{Problem Formulation} \label{sec:problem_formulation}
We study three tasks around creative refinement as described below.
\subsection{Task 1: ad text generation} \label{sec:task1}
In this task, the goal is to generate refined ad text (output) given an input ad (text and image).
For example, considering the illustration in Figure~\ref{fig:pull_figure} for advertiser $\#N$, if the input ad text is `\textit{great offers on cowboy boots!}', a possible generated output could be
`\textit{limited time offer on new cowboy boots!}'.
We assume that the input ad image is retained for use with the output ad text.
Additional metadata in the form of ad image (tags) and associated advertiser category is also assumed to be available. The output ad text is expected to have at least $\Delta \%$ better CTR performance compared to the input ad text (where $\Delta$ is a design choice) and the output ad text is assumed to be targeted to the same population of users as the input ad.
\subsection{Task 2: ad text keyphrase ranking}\label{sec:task2}
This is a simpler variant of task $1$, where instead of generating the entire ad text, the task is to recommend keyphrases in the refined ad text. We formulate this as a ranking problem,
where one needs to rank keyphrases from a given vocabulary, for inclusion in the refined ad text. For example, in Figure~\ref{fig:pull_figure} for advertiser $\#N$, if the input ad text is `\textit{great offers on cowboy boots!}', a recommended list of keyphrases could have `\textit{limited time}' and `\textit{new}' as the top ranked keyphrases.
The motivation here is to study cases when target text generation is hard to achieve, but useful keyphrase recommendations can still be provided.
The objective is to recommend keyphrases that would increase the CTR if included in the ad text while keeping all other aspects of the ad (such as ad image) constant.

\subsection{Task 3: ad image tag ranking}
\label{sec:task3}
In this task, given an input ad image and text,
the goal is to recommend image tags (output) to refine the ad image. Image tags essentially correspond to objects in the image, and are sourced from a given vocabulary of tags (explained later in Section~\ref{sec:kp_extraction}). This task is the visual parallel of task $2$, where instead of recommending textual keyphrases, we recommend tags for refining the ad image. The image tags can be used to select an ad image from a pool of images (\emph{e.g.}, via a stock image library \cite{shutterstock}); however, selecting or generating the final ad image is beyond the scope of this paper, and our study is limited to recommending image tags for the refined ad image.
For example, in Figure~\ref{fig:pull_figure} for advertiser $\#1$, if the input ad text is the one with multimedia devices, a recommended list of image tags could contain `\textit{human}' as a top ranked image tag. In addition, we assume that the input ad text is retained, and is available as metadata along with the associated advertiser category. The selection of an ad image based on the recommended tags is expected to increase the CTR of the refined creative.

\section{Data} \label{sec:data}
In this section, we first explain the ad platform setup in Section~\ref{sec:platform_setup}; specifically Yahoo Gemini ad platform, however, the underlying hierarchical structure is fairly standard in the advertising industry. This is followed by our method for leveraging the ad platform setup to form ordered pairs of creatives (Section~\ref{sec:pairs_generation}); the ordered pairs of creatives have a crucial role in our proposed methods to solve the tasks outlined in Section~\ref{sec:problem_formulation}. In Section~\ref{sec:kp_extraction}, we cover additional steps to automatically annotate the ad creative pairs with matched keyphrases and identified image tags. Finally in Section~\ref{sec:insights}, we describe data insights which motivate our approaches.
\begin{figure}[!htb]
\centering
  \includegraphics[width=0.7\columnwidth]{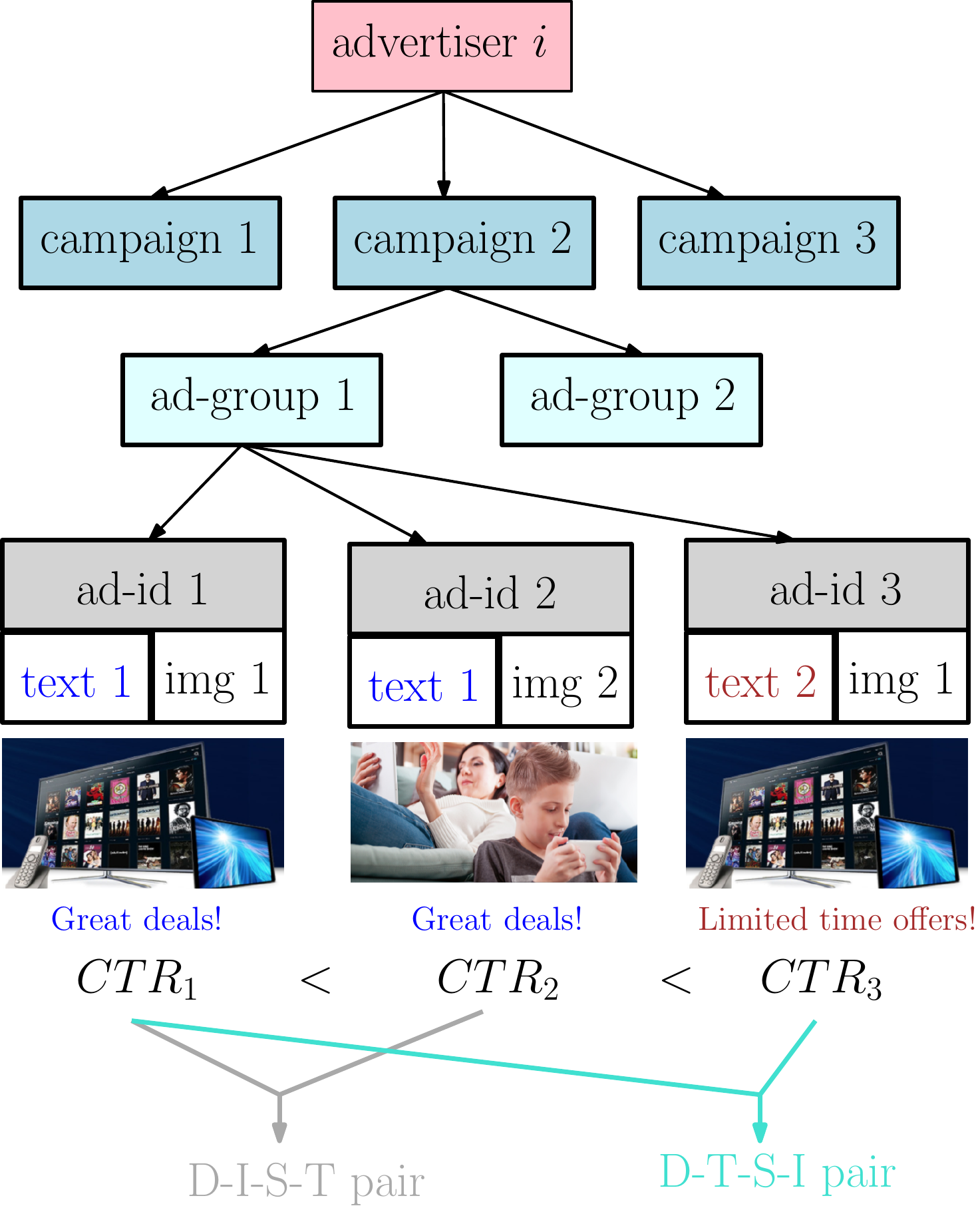}
  \caption{Ad campaign setup with multiple ad-groups and ad-ids. Difference in CTRs across ad-ids in the same ad-group can be attributed to differences in ad text and image. Ad-ids $1$ and $2$ form a different-image-same-text (D-I-S-T) pair, while $1$ and $3$ form a different-text-same-image (D-T-S-I) pair.}
  \label{fig:cmp_setup}
\end{figure}

\subsection{Ad platform setup} \label{sec:platform_setup}
As shown in Figure~\ref{fig:cmp_setup}, an advertiser in the Yahoo Gemini ad platform can create multiple campaigns and each campaign can have multiple ad-groups. Each ad-group is tied to a pre-specified target audience. For example, if the advertiser is a major telecommunications company, different campaigns may represent different offerings from the company (\emph{e.g.}, mobile phone plans and WiFi routers) whereas examples of ad-group targeting can be \textit{seniors in New York City} and \textit{males in San Francisco}. As shown in Figure~\ref{fig:cmp_setup}, there can be multiple ad-ids in an ad-group; each ad-id has an ad text and image associated with it.
For each qualifying user for the ad-group, one of the ad-ids is shown at random; in other words, if there is CTR performance difference across the ad-ids, it can be purely attributed to the differences in ad image and text across the ad-ids in the ad-group. For the example shown, the difference in CTRs of ad-id $1$ and $2$ can be attributed to the difference in the ad image, while for ad-ids $1$ and $3$, the difference can be attributed to the difference in ad text. However, in the case of ad-ids $2$ and $3$, the CTR difference is a result of differences in both the image and text.
\subsection{Constructing ad creative pairs}\label{sec:pairs_generation}
We use data from ad-groups across multiple advertisers to form two datasets: (i) different-text-same-image (D-T-S-I) dataset,
and (ii) different-image-same-text (D-I-S-T) dataset as described below.
\subsubsection{D-T-S-I dataset} \label{sec:DTSI}
To create this dataset, from each ad-group, we create
pairs of ad-ids (creatives) such that in each pair the ad text is different but the ad image is same. Furthermore, in each such pair, we order the ad-ids as (source, target) where source CTR is lower than target CTR. For example, in Figure~\ref{fig:cmp_setup}, (ad-id $1$, ad-id $3$) form such a (source, target) pair in the D-T-S-I dataset. We collect such pairs using ad-groups across multiple advertisers. In case multiple pairs have the same source ad text (but different target ad text), we only retain the pair with highest CTR difference, and discard the other (duplicate-source) pairs. Finally, we keep the pairs where the relative CTR difference is higher than $\Delta\%$ (design choice). The intuition behind creating such pairs is to provide training examples to an ad text refinement model, \emph{e.g.}, for generating the target ad text given the source ad text (explained in Section~\ref{sec:generation}).

\subsubsection{D-I-S-T dataset} \label{sec:DIST}
To create this dataset, from each ad-group, we create pairs of ad-ids such that the ad image is different but the ad text is same. As in the D-T-S-I dataset, we order the ad-ids in the pair as (source, target) where source CTR is lower than target CTR; in Figure~\ref{fig:cmp_setup}, (ad-id  $1$, ad-id $2$) is an example of such a pair. We collect such pairs across ad-groups of multiple advertisers. If there are pairs with the same source image, we retain the pair with the highest CTR difference and discard the other duplicates. Finally, we filter out pairs with relative CTR difference below $\Delta\%$ (design choice). The intuition behind creating such pairs is to provide training examples for refined ad images given source ad text and image.

\subsection{Keyphrases and image tags annotation}\label{sec:kp_extraction}
For each pair in the D-T-S-I and D-I-S-T datasets, we add metadata in the form of matched keyphrases and image tags (explained below).
\paragraph*{Keyphrases} We first form a vocabulary of keyphrases using an unsupervised keyphrase extraction method\footnote{We used multipartite-rank \cite{multipartite_rank} method implemented in the PKE keyphrase extraction package \cite{pke}. Choice of this method (versus others in PKE, \emph{e.g.}, TF-IDF, and Position-rank \cite{position-rank}) was guided by visual inspection of results on representative advertisers.} on the collective ad text corpus (including both source and target ad text from all pairs). For example, from retail advertisers, typical examples of extracted keyphrases include phrases like \textit{free shipping} and \textit{limited time offers}, while from telecommunication advertisers, examples include \textit{high speed internet} and \textit{bundle deals}. Using the obtained vocabulary of keyphrases, for each pair in the D-T-S-I and D-S-T-I datasets, we add a list of exact matches found in the source and target ad text.

\paragraph*{Image tags}
Image tags are the objects detected in an image via the (pre-trained) Inception Resnet v2 object detection model as in the Open Images V2 repository \cite{openimages}. We extract these image tags from the source and target ad images in D-T-S-I and D-I-S-T datasets. 
Inception Resnet v2 \cite{openimages} is a convolutional neural network trained by Google on Flickr images in the Open Images V2 dataset. It has about $5000$ classes (possible tags in an image). Each image can have multiple tags and the model returns a list of inferred tags with confidence scores. We retain all tags with a score above $0.8$. For example, the ad image in ad-id $2$ in Figure~\ref{fig:cmp_setup} has tags \textit{woman, child, face}, whereas the image in ad-id $1$ has the tag \textit{multimedia}.

\subsection{Insights from D-I-S-T and D-T-S-I datasets}\label{sec:insights}
Based on $5$ months (July--November 2019) of data from the Yahoo Gemini platform, we gathered several insights from D-T-S-I and D-I-S-T datasets spanning a sample of over $3500$ advertisers. The minimum CTR difference ($\Delta$) in each source-target pair was kept at $10\%$. We highlight key insights below which guided our proposed approaches
(additional statistics are covered later in Section~\ref{sec:results}).

\paragraph*{High word overlap between source and target text}
In the D-T-S-I dataset, the average number of words in both target and source ad text is close to $13$ (sequence length), but there is a $60\%$ overlap between words in source and target. This indicates: (i) target retains a lot of words from the source (plausibly to preserve context), and (ii) there are word replacements in source to keep the sequence length roughly the same.
Hence, a \textit{copy mechanism} which can selectively copy parts of the source text while introducing new words in target looks intuitive for the ad text generation task  (details in Section~\ref{sec:generation}).

\paragraph*{Discriminative power of keyphrases and image tags}
An advertiser category-wise case study using the D-T-S-I dataset revealed that the presence of certain keyphrases in the target ad text (and their absence in the source) consistently led to higher CTR relative to the source. For example, in the case of retail category advertisers, such keyphrases included \textit{free shipping} and \textit{limited time offer}. In a parallel study using the D-I-S-T dataset, we observed analogous results with image tags. For example, for telecommunication advertisers, we found that target images with human elements (\emph{i.e.} having tags \textit{woman}, \textit{man}, \textit{child}) had higher CTR than source images with just \textit{multimedia} tag.
The above insights motivate the use of a ranking approach for recommending keyphrases and image tags for refining an input ad creative (details in Section~\ref{sec:ranking}).

\section{Generation and ranking models} \label{sec:method}
We now describe, the proposed solutions for tasks 1-3 (Section~\ref{sec:problem_formulation}). The text generation approach for task $1$ (Section~\ref{sec:task1}) is explained in Section~\ref{sec:generation}.
For tasks $2$ (Section~\ref{sec:task2}) and $3$ (Section~\ref{sec:task3}), the proposed keyphrase/image tag ranking model is explained in Section~\ref{sec:ranking}.

\subsection{Ad text generation model for task $1$} \label{sec:generation}
Task $1$ can be formulated a sequence-to-sequence (seq2seq) prediction task, where given an input ad text (source sequence of tokens), the predicted output should be a refined version of the input ad text (target sequence) with a higher expected CTR. The construction of the D-T-S-I dataset (in Section~\ref{sec:DTSI}) is naturally suited for training such a seq2seq model, since in each pair the target ad text has higher CTR than the source ad text (the same ad-group, and same image constraints in each pair eliminate all other confounding factors affecting CTR). Given the D-T-S-I dataset, to solve task $1$, we propose using an encoder-decoder architecture with a mechanism to selectively \textit{copy} words from the source text; the intuition behind the proposal, and underlying architecture details are explained below.

\begin{figure}[!htb]
\centering
  \includegraphics[width=0.8\columnwidth]{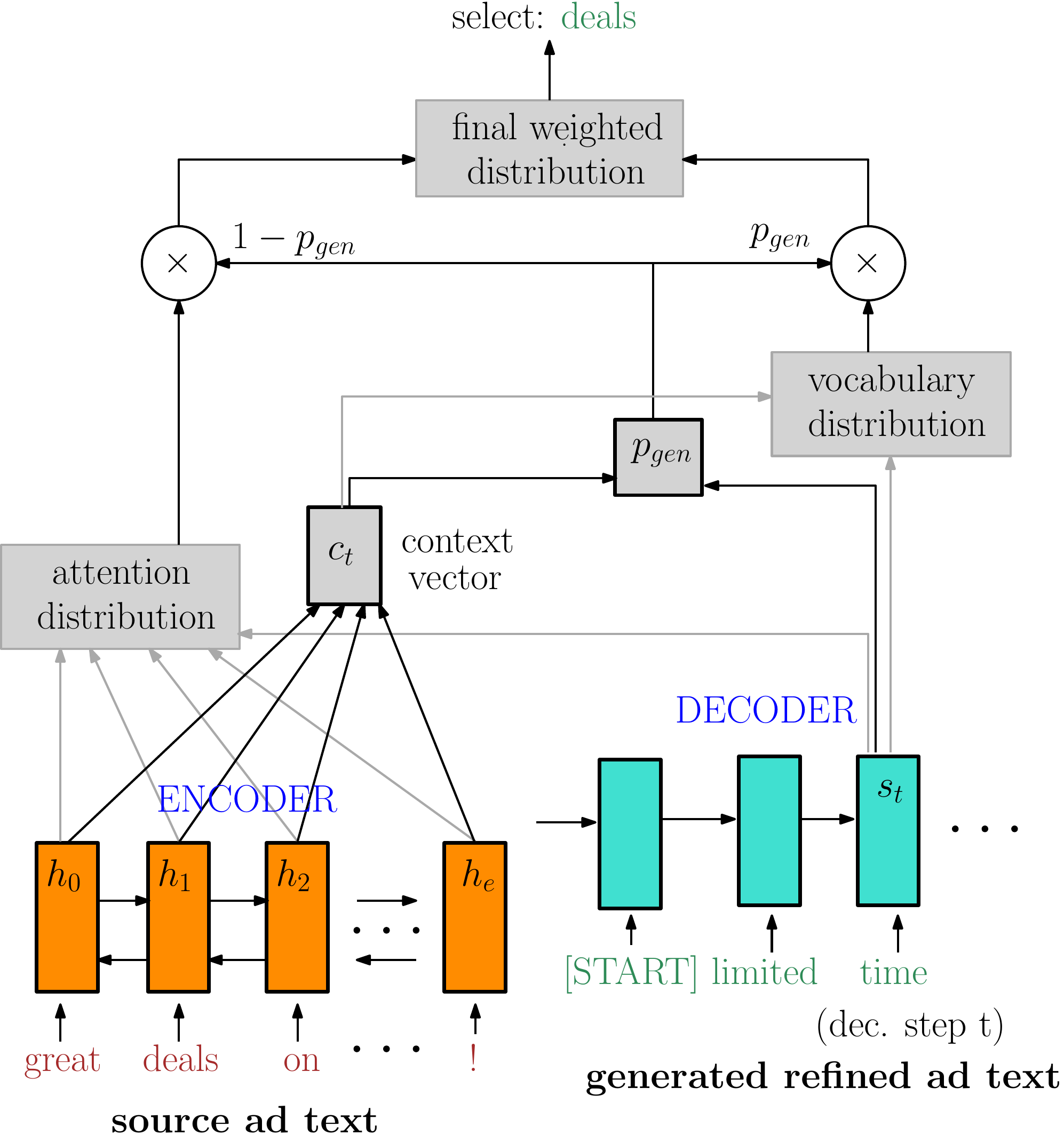}
  \caption{Encoder-decoder with attention and copy mechanism for generating refined (target) ad text given source ad.}
  \label{fig:copy_mech}
\end{figure}

\paragraph*{Intuition} We borrow ideas from state-of-the-art models in abstractive summarization \cite{see_pointer_generator} and use it to solve task $1$ as follows. We use an encoder-decoder architecture with attention \cite{bahdanau2014neural}, along with a copy mechanism \cite{see_pointer_generator} as shown in Figure~\ref{fig:copy_mech}.
In our setup, the motivation for using the copy mechanism is driven by the observation that there is a $60\%$ overlap between source and target words in the D-T-S-I dataset (as mentioned in Section~\ref{sec:insights}). It is plausible that copying some words from the source is good enough to preserve the underlying context, while adding new words in the target can boost the CTR. We describe the underlying model details below.

\paragraph*{Model details:}
We use a bidirectional LSTM encoder for the source sequence
and an LSTM decoder for the target sequence \cite{bahdanau2014neural}.
Following \cite{luong2015attention}, the attention distribution is computed as:
\begin{align}
e^t_i &= h_i^\top W_\text{att} s_t,
\quad a^t = \text{softmax}\left( e^t \right),
\end{align}
where $h_i$ is the encoder hidden state, $s_t$ is decoder state at step $t$, $a^t$ is the attention distribution, and $W_\text{att}$ represents the learnable parameters.
The attention-weighted sum of all encoder hidden states is used compute the context vector as:
\begin{align}
    c_t = \sum_i a^t_i h_i.
\end{align}
%
The generation probability $p_{gen}$ for step $t$ is computed using the context vector ($c_t$), decoder state ($s_t$) and decoder input ($x_t$) as:
\begin{align}
    p_\text{gen} = \sigma\left( w^\top_c c_t + w_s^\top s_t + w_x^\top x_t + b_\text{ptr}\right),
\end{align}
where $w_c$, $w_s$, $w_x$ are vectors and
$b_\text{ptr}$ is a scalar, all of which are learnable; $\sigma(\cdot)$ denotes the sigmoid function. Here, $p_\text{gen}$ is used to \textit{softly} choose between generating a word from the entire vocabulary versus copying a word (token) from the input sequence (via sampling from the attention distribution $a^t$). The vocabulary distribution for generating a new word can be computed as:
\begin{align}
    \mathbb{P}_\text{vocab} = \text{softmax}\left(
    V' ( V \left [ s_t; c_t \right ] + b ) + b' \right),
\end{align}
where $V$, $V'$, $b$, and $b'$ are learnable parameters. With $p_{gen}$,
the effective distribution over the vocabulary can be written as:
\begin{align}
\mathbb{P}(y) = p_\text{gen} \mathbb{P}_\text{vocab}(y) 
+ (1-p_\text{gen}) \sum_{i:y_i=y} a^t_i,
\end{align}
where $y$ is a word in the vocabulary.
For training, the loss at step $t$ ($\mathcal{L}_t$) is the negative log-likelihood associated with target word $y^\ast_t$, and that of the whole sequence is simply the average:
\begin{align}
\mathcal{L}_{t} = -\log \left(  \mathbb{P}(y^\ast_t)\right), \quad 
\mathcal{L} = \frac{1}{T} \sum_{t=0}^T \mathcal{L}_{t}.
\end{align}
Our implementation of the above model leveraged OpenNMT-Py \cite{opennmt} with: train steps = 200k, optimizer = SGD, and batch size = 128.
\subsection{Ranking model for tasks $2$ and $3$} \label{sec:ranking}
\begin{figure}[!htb]
\centering
  \includegraphics[width=0.7\columnwidth]{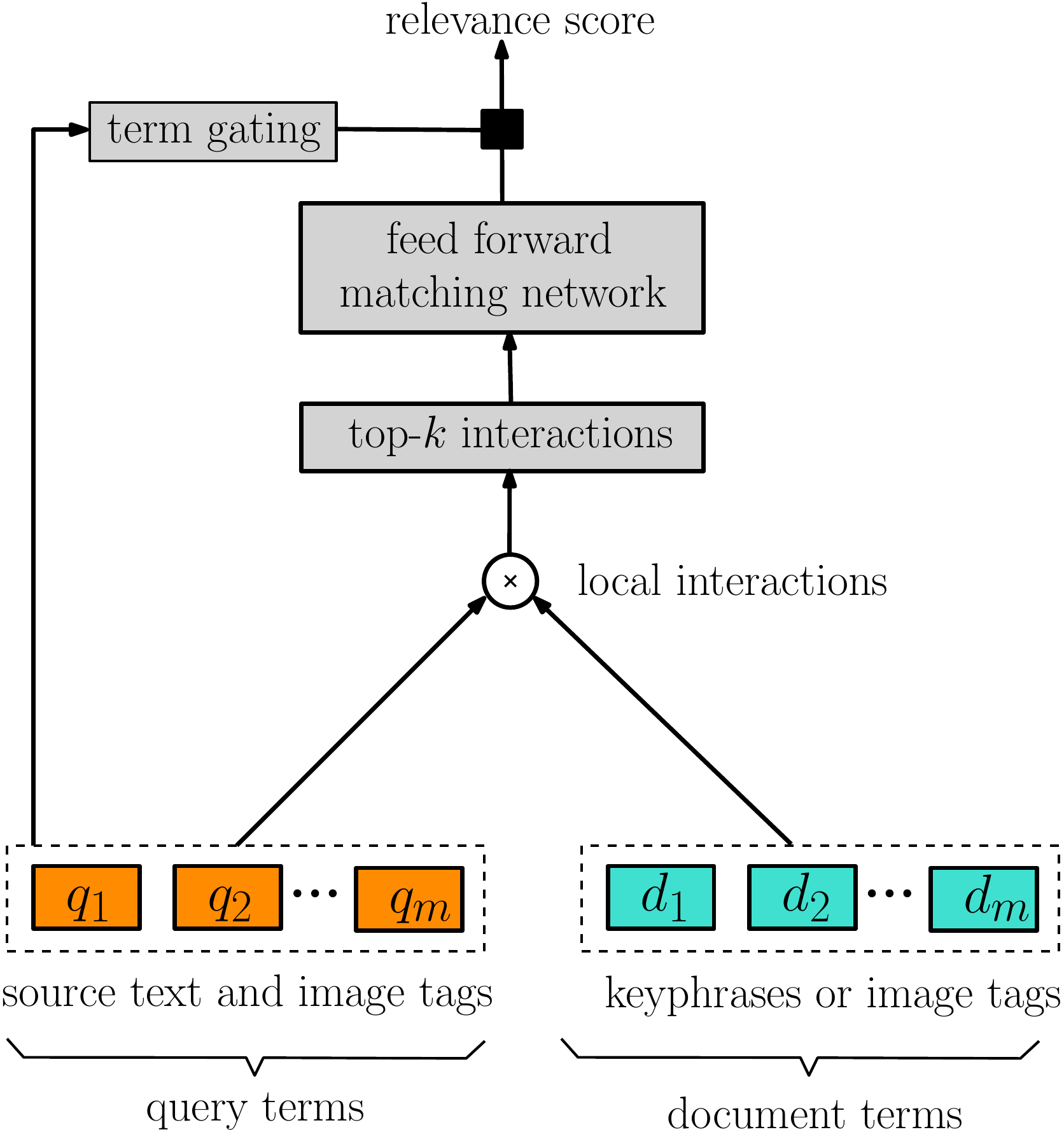}
  \caption{DRMM-(top $k$) for keyphrase/image tag ranking.}
  \label{fig:drmm}
\end{figure}

We consider solving the keyphrase (and image tag) recommendation problem via a ranking model, where the model outputs a list of keyphrases (and image tags) in decreasing order of relevance for a given ad creative.
We describe below the model for the keyphrase ranking task; the image tag ranking model is analogous, and we skip its description for brevity.
We use the state-of-the-art pairwise deep relevance matching model (DRMM)~\cite{drmm,drmm_topk} whose architecture for our recommendation setup is shown in  Figure~\ref{fig:drmm}. It is worth noting that our pairwise ranking formulation can be changed to accommodate other multi-objective or list-based loss-functions. We chose the DRMM model since it is not restricted by the length of input, as most ranking models are, but relies on capturing local interactions between query and document terms.
Given a \emph{(source ad text , target keyphrase)} combination, the model first computes the top-$k$ interactions between the source ad text words and the keyphrases. These interactions are passed through a multi-layer perceptron (MLP), and the overall score is aggregated with a query term gate which is a softmax function over all terms in that query.
DRMM employs a pair-wise ranking loss function as described below. We denote the source ad text by just $src$ in the following explanation. Given a triple ($src$, $p^+$, $p^-$) where keyphrase $p^+$ is ranked higher than keyphrase $p^-$ with respect to $src$, the loss function is:
\begin{align}
\mathcal{L} (src,p^+,p^-;\theta)=max(0, 1-s(src,p^+) + s(src,p^-)) ,
\end{align}
where $s(src, p)$ denotes the predicted matching score for keyphrase $p$, and the source ad text. Metadata in the form of image tags,
and  advertiser category can be introduced as additional query terms.
In our implementation, we used the top-k version of DRMM \cite{drmm_topk} in Match-Zoo \cite{matchzoo} with $k=20$ and ADAM optimizer.

\section{Results} \label{sec:results}
In this section, we first cover notable statistics of the D-T-S-I and D-I-S-T datasets in Section~\ref{sec:stats}, followed by a description of evaluation metrics in Section~\ref{sec:metrics}. This is followed by results on ad text generation, keyphrase ranking, and image tag ranking.

\begin{table}[th]
    \centering
    \begin{tabular}{|l||l|l||l|l|}
    \hline 
       task     &\multicolumn{2}{|l|}{ keyphrase ranking} &\multicolumn{2}{|l|}{ image-tag ranking} \\ \hline
             & source  & target  &  source  & target \\ \hline
    \multicolumn{5}{|l|}{ vanilla-split }    \\ \hline
     \# tokens (1)   & 12.24$\pm3.6$ & 12.25$\pm3.6$ & 13.44$\pm3.8$ & 13.44$\pm3.8$ \\
     \# tokens (2)    &  12.37$\pm3.6$ & 12.32$\pm3.6$ & 13.44$\pm3.7$ &  13.44$\pm3.7$ \\
     \# kp/img (1)  & & 12.02$\pm4$  & & 7.14$\pm3.8$ \\ 
     \# kp/img (2)  & &12.13$\pm4$ & & 7.06 $\pm3.8$ \\ 
     \hline
     \end{tabular}
    \caption{Mean ($\pm$std) of attributes of train (1) and test (2) sets: number of words in ad text (\# tokens), number of matched keyphrases (\# kp), and number of image tags (\# img).}
    \label{table:dataset_stats}
\end{table}

\begin{table*}[h]
    \centering
    \begin{tabular}{|l|c|c|c|c|c|c|c|}
    \hline 
     model &  BLEU & ROUGE-1 F & ROUGE-2 F & ROUGE-L F & kp-P & kp-R & kp-F \\
     \hline 
     \hline
     \textbf{vanilla-split}  &  &   &  &  &  &  &       \\ 
    \hline
     \hline 
    baseline (pred=src) & 56.28	&	63.49	&	50.79	&	61.13	&	0.643	&	0.644	&	0.643   \\ 
    ATTN    & 50.74	&	57.62	&	47.26	&	56.01	&	0.552	&	0.548	&	0.55     \\ 
    ATTN + CP  & 59.38	&	65.61	&	55.13	&	63.79	&	0.661	&	0.648	&	0.655      \\
    ATTN + CP + CAT & \textbf{59.45}	&	\textbf{65.74}	&	\textbf{55.35}	&	\textbf{63.91}	&	\textbf{0.661}	&	\textbf{0.649}	&	\textbf{0.655}  \\ 
    ATTN + CP + CAT + IMG & 58.37	&	65.63	&	55.18	&	63.82	&	0.663	&	0.646	&	0.654    \\ 
    \hline
    \hline
    \textbf{cold-start split}  &  &   &  &  &  &  &       \\ 
    \hline
    \hline
    baseline (pred=src)       & 56.01	&	63.69	&	51.02	&	61.57	&	0.643	&	0.637	&	0.64      \\
    ATTN             &16	&	26.64	&	13.29	&	25.02 &	0.195 &	0.177 &	0.185 \\
    ATTN + CP        & 34.39	&	45.26	&	30.86	&	42.81	& 0.462	& 0.422 &	0.441	\\
    ATTN + CP + CAT  & \textbf{35.91}	&	\textbf{47.52}	&	\textbf{32.64}	&	\textbf{44.69}	& \textbf{0.494} &	\textbf{0.434} &	\textbf{0.462}	   \\
    ATTN + CP + CAT + IMG & 33.42	& 44.33		&	29.53	&  41.76 & 0.422 &	0.37 &	0.394  \\
    \hline
    \end{tabular}
    \caption{Ad text generation results: ATTN denotes the LSTM encoder-decoder with attention model, CP denotes copy mechanism, CAT denotes adding category, and IMG denotes adding source image tags.}
    \label{tab:generation_results}
\end{table*}

\subsection{Dataset statistics} \label{sec:stats}
The D-T-S-I and D-I-S-T datasets were built using a sample of $5$ months of data from Yahoo Gemini (July-November 2019). The data consisted of over $3500$ advertisers ($> 8500$ campaigns in English for U.S. audiences, $\sim 100$ categories), and each ad-id considered in the dataset had over $10,000$ impressions. After the filtering process (\emph{i.e.}, keeping only source target pairs with more than $\Delta=10\%$ CTR difference, and removing duplicate sources), the D-T-S-I dataset consisted of over $20,000$ pairs while the D-I-S-T dataset consisted of over $10,000$ pairs.
Each dataset was randomly divided into train, test and validation sets in proportions of $77\%$, $20\%$, and $3\%$ respectively; we will refer to this as a \textit{vanilla} split. In addition to the vanilla split, we created a cold-start split where there was no overlap between advertisers in train, test and validation sets; 
this presents a much more difficult (versus vanilla split) learning problem with unseen advertisers
Additional dataset statistics are shown in Table \ref{table:dataset_stats} .

\subsection{Evaluation metrics} \label{sec:metrics}
For ad text generation, we use standard metrics for text generation problems: (i) BLEU \cite{BLEU}, and (ii) ROUGE scores \cite{ROUGE}. We introduce metrics to gauge the presence of matched (target) keyphrases in the generated sequence: (i) keyphrase-precision (kp-P),
(ii) keyphrase-recall (kp-R), and (iii) keyphrase -F (kp-F).
In other words, we compute precision and recall for target keyphrases, considering the list of tokens in the generated text.
For both keyphrase and image tag ranking,
we use:
(i) precision at $k$ ($P@k$),
(ii) recall at $k$ ($R@k$), and (iii) normalized cumulative discounted gain at $k$ ($NDCG@k$).


\subsection{Ad text generation results}
Table ~\ref{tab:generation_results} covers generation results for the vanilla, and cold-start cases (metrics on test set).
The baseline scores are for the case when the source ad text is considered as the predicted ad text (\emph{i.e.}, no change in input), and compared with the target ad text. In Table~\ref{tab:generation_results}, CAT and IMG denote the addition of category and image tags to beginning of the input sequence (image tags in alphabetical order). The main observations are as follows.
\paragraph*{Copy mechanism works} In both vanilla and cold-start cases, there is a significant lift in the metrics due to the copy mechanism. In case of vanilla split, the copy mechanism is able to beat the baseline (predicted sequence = source sequence) metrics. However, in cold-start, it is below the baseline (but is better than the no-copy version).

\paragraph*{Category helps} There is a consistent improvement in metrics on using category metadata in the input sequence. As expected, category information provides a relatively higher lift ($4.4\%$ above ATTN+CP in ROUGE-L F) for cold-start split compared to vanilla split ($0.2\%$ lift). In comparison, adding image tags to the input sequence (along with category) does not provide any lift (suggesting the need for better ways to incorporate image information).

\begin{figure}[]
\centering
  \includegraphics[width=1 \columnwidth]{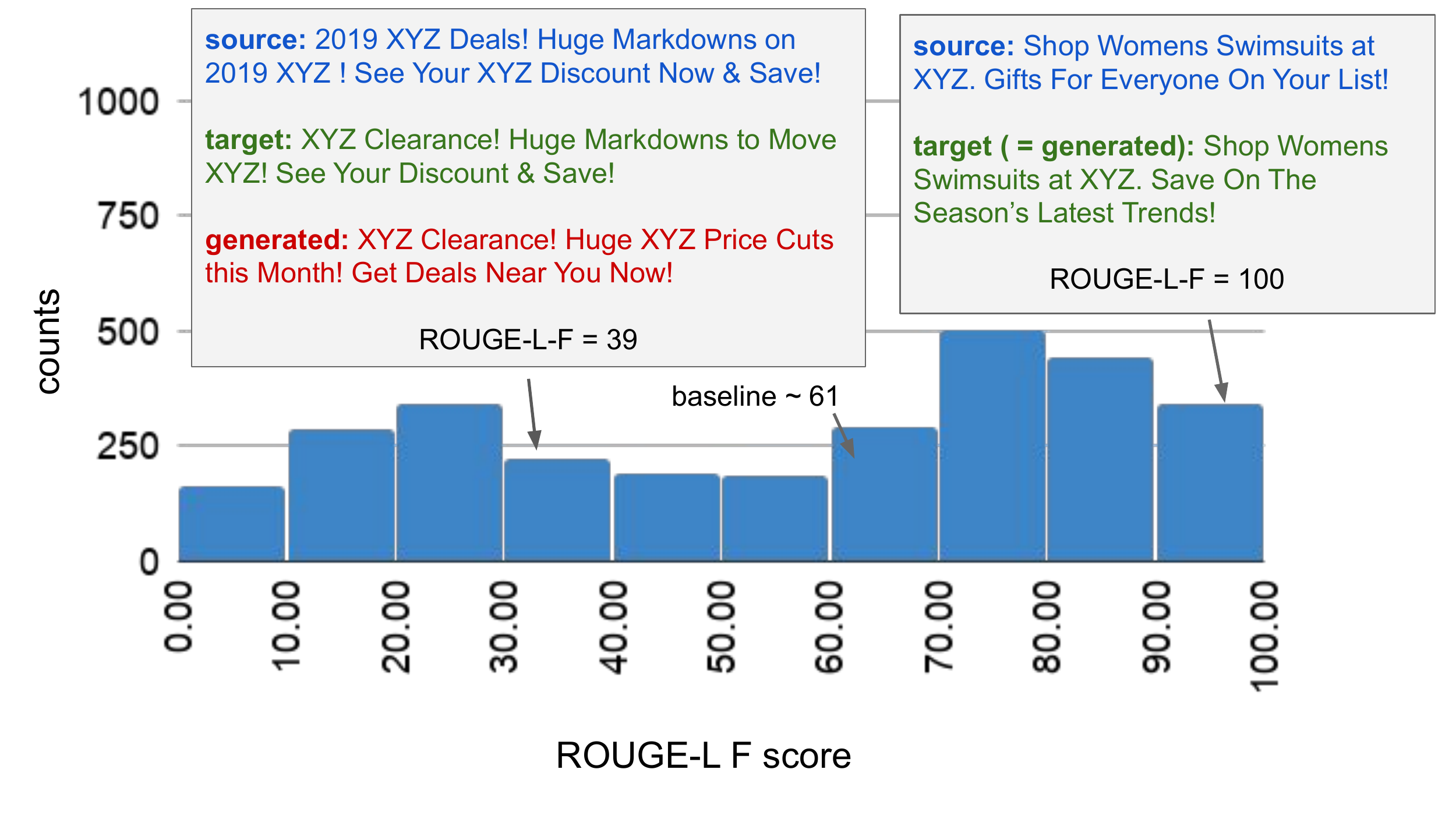}
  \caption{Histogram of ROUGE-L F scores in test set for vanilla-split (ATTN + CP + CAT model). Two anonymized examples are also shown with their ROUGE-L F scores.}
  \label{fig:hist_vanilla}
\end{figure}

\begin{figure}[]
\centering
  \includegraphics[width=1 \columnwidth]{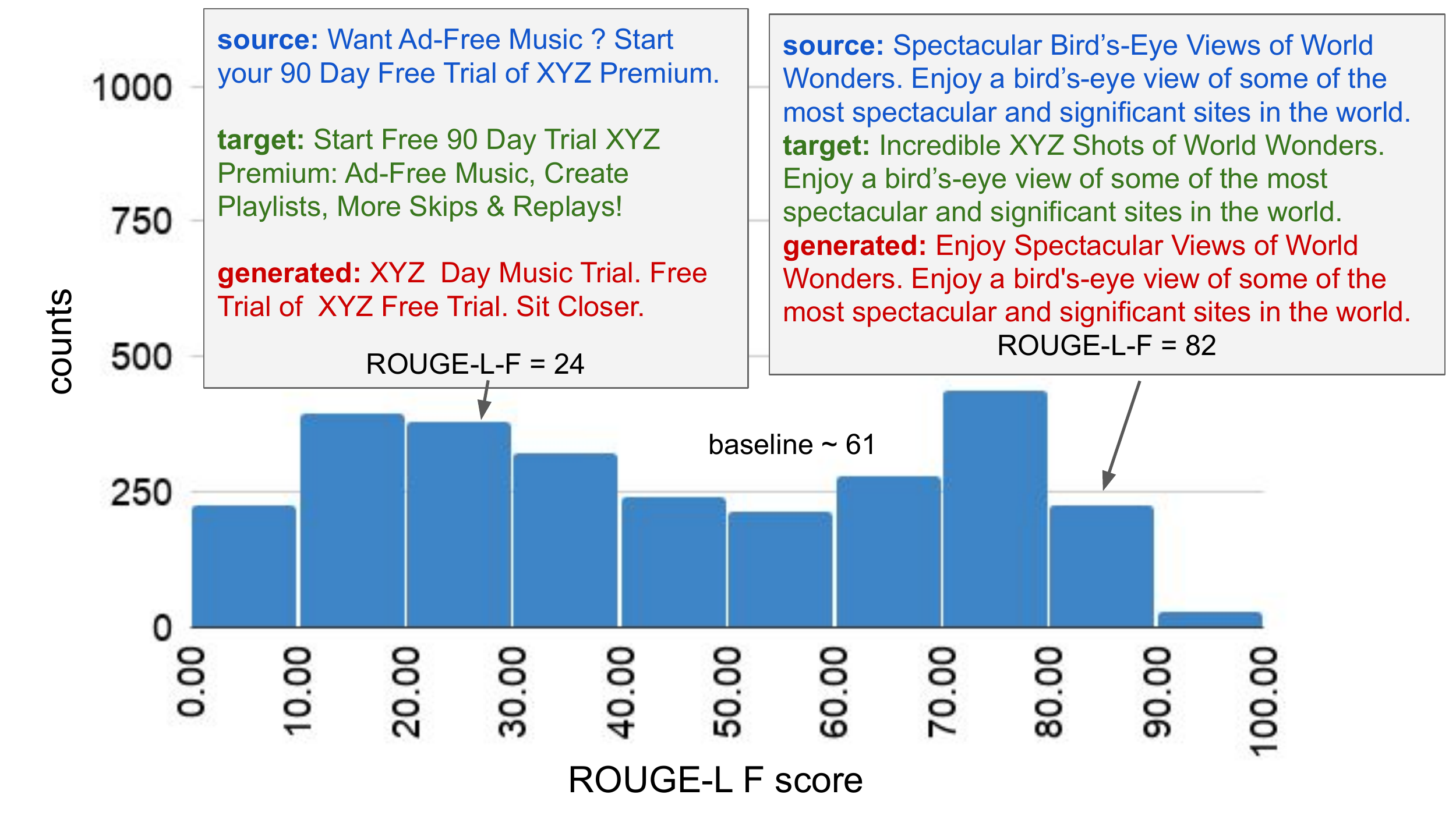}
  \caption{Histogram of ROUGE-L F scores in test set for cold-start split (ATTN + CP + CAT model). Two anonymized examples are also shown with their ROUGE-L F scores.}
  \label{fig:hist_coldstart}
\end{figure}

\paragraph*{Cold-start is challenging}
We computed the histogram of ROUGE-L F scores on the test set using the best model (ATTN+CAT+CP) for the vanilla-split (Figure~\ref{fig:hist_vanilla}) and cold-start split (Figure~\ref{fig:hist_coldstart}) cases; for both splits, the baseline ROUGE-L F is around $61$. As shown, for vanilla, the distribution has a significant number examples above the baseline, while the distribution's mass significantly shifts below baseline for cold-start. The listed examples of generated text give a sense of how good the generated outputs are in terms of human judgement vis-a-vis ROUGE-L F scores. 
The keyphrase based metrics for generation (kp-P,R,F) as reported in Table~\ref{tab:generation_results}, are helpful in gauging the extent to which target keyphrases appear in the generated text. For example,
in the lower scored generated text in Figure~\ref{fig:hist_vanilla}, \textit{clearance} is correctly introduced, but \textit{price cuts} is incorrectly introduced. Although \textit{price cuts} is incorrect given the target text (which has a guaranteed CTR lift), it remains to be seen if it leads to a lower CTR online (beyond the scope of this paper).
\subsection{Keyphrase ranking results} \label{sec:keyphrase_ranking}
\begin{figure}[]
\centering
  \includegraphics[width=1 \columnwidth]{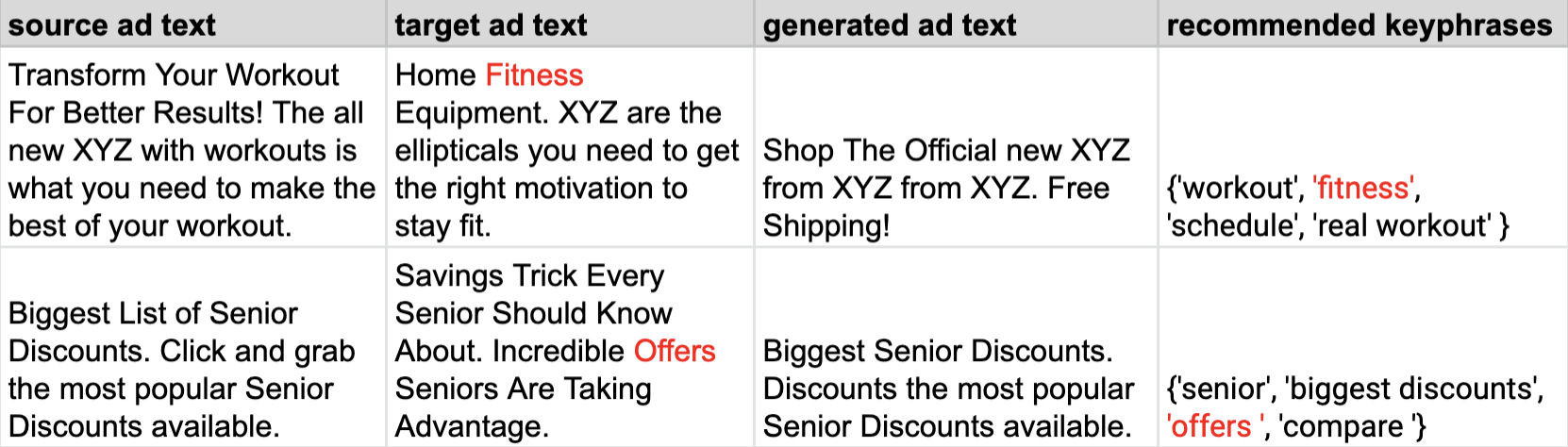}
  \caption{Sample keyphrase ranking results vis-a-vis generated ad text (cold-start); the keyphrase recommendations can cover target keyphrases missed by poor generation.}
  \label{fig:ranking_examples}
\end{figure}

Table~\ref{tab:ranking_results_text} shows the results for keyphrase ranking.
The baselines included methods using:
(i) cosine similarity (EMB-SIM) based on Glove \cite{glove} embeddings for keyphrases and input text (average of word embeddings), and 
(ii) TF-IDF representation of source ad text and keyphrases is used to compute similarity and keyphrases are ranked in descending order of similarity.
\begin{table}[h]
    \centering
    \begin{tabular}{|l|c|c|c|c|c|c|}
    \hline 
     model &  P\@5 & P\@10 & R\@5 & R\@10 & ndcg\@5 & ndcg\@10 \\
     \hline 
     \hline
     \textbf{vanilla}  &   & & & & &  \\ 
    \hline
     \hline 
    EMB-SIM	&	0.17	&	0.10	&	0.07	&	0.09	&	0.19	&	0.14 \\
    TF-IDF	&	0.33	&	0.26	&	0.15	&	0.23	&	0.35	&	0.30 \\ 
    DRMM	&	0.50	&	0.39	&	0.25	&	0.38	&	0.53	&	0.47 \\
    \hline
    + CAT 	    & \textbf{0.51}	&	\textbf{0.40}	&	\textbf{0.25}	&	\textbf{0.39}	& \textbf{0.53}	& \textbf{0.48}	 \\
    + CAT + IMG	& 0.41	&	0.32	&	0.21	&	0.32	&0.43	&0.39	 \\
    \hline
    \hline
    \textbf{cold st.}   &   & & & & &    \\
    \hline
    \hline
    EMB-SIM	&	0.12	&	0.07	&	0.05	&	0.06	&	0.14	&	0.11 \\
    TF-IDF	&	0.27	&	0.21	&	0.12	&	0.18	&	0.29	&	0.26	\\
    DRMM	&	0.38	&	0.29	&	0.22	&	0.32	&	0.41	&	0.37	\\ 
    \hline
    + CAT 	    & \textbf{0.42}	&	\textbf{0.32}	&	\textbf{0.24}	&	\textbf{0.36}	&	\textbf{0.45}	&	\textbf{0.40}	 \\
    + CAT + IMG	&	0.34	&	0.26	&	0.20	&	0.30	&	0.36	&	0.33	\\
    \hline
    \end{tabular}
    \caption{Kephrase ranking: baselines versus DRMM, and the effect of adding category and image tags as query terms.}
    \label{tab:ranking_results_text}
\end{table}
As shown in Table~\ref{tab:ranking_results_text}, for both splits, using DRMM with category features performs the best in terms of all metrics.
\begin{table}[h]
    \centering
    \begin{tabular}{|l|c|c|c|c|c|c|c|}
    \hline 
     split	&	metric	&	add-0	&	add-1	&	add-2	&	add-3	&	add-10 \\
     \hline
\hline
cold-start	&	kp-P	&	0.50	&	0.50	&	0.49	&	0.46	&	0.35 \\
cold-start	&	kp-R	&	0.43	&	0.45	&	0.46	&	0.47	&	0.53 \\
     \hline
     \end{tabular}
    \caption{Ranking-aided keyphrase metrics for generation. Add-0 denotes no assistance, and add-10 denotes adding top $10$ ranked keyphrases in the generation output.}
    \label{tab:assisted_kp}
\end{table}
Cold-start best performance is comparable to the vanilla split best performance (\emph{e.g.}, $7\%$ drop in $R@10$, compared to $33\%$ drop in $kp-R$ in Table~\ref{tab:generation_results} for generation). Hence, keyphrase ranking seems to be more \textit{robust} to unseen advertisers compared to ad text generation. As seen in text generation, naively adding image tags to the input along with category does not generalize well (mildly hurts performance).
We suspect that since image tags represent objects, they provide no additional context for the ranker to select better keyphrases. Most often, keyphrases provide more information about the brand, and image tags that represent objects may not add any complementary information about the brand directly that the ranking model can exploit. In future, we shall explore features that encapsulate information in the image directly \cite{www20_joey} rather than use image tags for keyphrase ranking.
We also study the possibility of \textit{assisting} generation results with corresponding ranking results. Table~\ref{tab:assisted_kp} shows the boost in kp-R for the best generation results (ATTN+CP+CAT), when the corresponding (top-$r$) outputs of the DRMM + CAT model are added to the list of matched keyphrases in generated text.
As shown for cold-start, just adding the top ranked keyphrase (add-1) improves the recall ($0.43\rightarrow 0.45$) without affecting the precision ($0.5$). This indicates that ranking results can complement generation results in a helpful manner (illustrative cold-start examples in Figure~\ref{fig:ranking_examples}).

\subsection{Ad image tag ranking results} \label{sec:image_tag_ranking}
Table~\ref{tab:ranking_results_image} shows the ranking results for image tags (baselines, and CAT + IMG feature additions in DRMM).
\begin{table}[h]
    \centering
    \begin{tabular}{|l|c|c|c|c|c|c|}
    \hline 
     model &  P\@5 & P\@10 & R\@5 & R\@10 & ndcg\@5 & ndcg\@10 \\
     \hline 
     \hline
     \textbf{vanilla}  &   & & & & &  \\ 
    \hline
     \hline 
    EMB-SIM	&	0.16	&	0.15	&	0.12	&	0.22	&	0.18	&	0.23 \\
    TF-IDF	&	0.27	&	0.24	&	0.21	&	0.35	&	0.29	&	0.37	\\
    DRMM	&	0.49	&	0.34	&	0.35	&	0.49	&	0.53	&	0.50	\\
    \hline
     + CAT	&	0.50	&	0.35	&	0.36	&	0.49	&	0.54	&	0.51	\\
     + CAT + IMG	&	\textbf{0.51}	&	\textbf{0.37}	&	\text{0.34}	&	\textbf{0.49}	&	\textbf{0.55}	&	\textbf{0.52}	\\
    \hline
    \hline
    \textbf{cold st.}   &   & & & & &    \\
    \hline
    \hline
    EMB-SIM	&	0.16	&	0.15	&	0.11	&	0.20	&	0.18	&	0.22	\\
TF-IDF	&	0.28	&	0.24	&	0.20	&	0.33	&	0.29	&	0.36	\\
DRMM	&	0.41	&	0.31	&	0.29	&	0.43	&	0.44	&	0.44	\\
\hline
 + CAT	&	0.43	&	0.33	&	0.31	&	0.45	&	0.46	&	0.46	\\
 + CAT + IMG	&	\textbf{0.53}	&	\textbf{0.37}	&	\textbf{0.38}	&	\textbf{0.52}	&	\textbf{0.58}	&	\textbf{0.55}	\\
\hline
    \end{tabular}
    \caption{Image tag ranking: baselines versus DRMM, and the effect of adding category and image tags as query terms.}
    \label{tab:ranking_results_image}
\end{table}
Using DRMM with category and image tags performs the best. The efficacy of image tags (in source) to predict relevant tags (in target) may be linked to common modality. \begin{table}[h]
    \centering
    \begin{tabular}{|l|c|}
    \hline 
     category & top $5$ ranked image tags  \\
     \hline
     apparel     & clothing, face, hair, girl, pattern \\
     job portals & face, clothing, multimedia, road, man, woman \\
     auto        & wheel, car, motorcycle, clothing, face \\
     real estate & man, woman, mansion, bedroom, kitchen \\ 
     \hline
    \end{tabular}
    \caption{Frequent top ranked image tags by category.}
    \label{tab:image_tag_insights}
\end{table}
We also report the frequent top ranked image tags for selected categories in Table~\ref{tab:image_tag_insights} using the DRMM + CAT + IMG model.

\subsection{Online results} \label{sec:online_results}
We deployed the ranking models for tasks $2$ and $3$ (\emph{i.e.}, keyphrase and image tag rankers) as an internal service for Yahoo Gemini account teams which manage campaigns of major advertisers. To study end-to-end adoption, we partnered with the account team for an Internet service provider. Using their existing creative (text and image) as input, the top keyphrase and top image tag recommendation were considered.
The advertiser approved an A/B test for the refined creative (incorporating both image and text refinements together) versus their existing creative. The A/B test was conducted for $2$ weeks via Yahoo Gemini, and the refined  creative showed an $87\%$ improvement in CTR, validating the model recommendations.
\section{Discussion} \label{sec:discussion}
Our results show the efficacy of using A/B test data across advertisers for both generation and ranking formulations of ad creative refinement.
Account teams testing the proposed models requested additional evidence in the form of CTR of similar ads (\emph{i.e.}, with recommended keyphrases and image tags) to convince advertisers to approve tests for refined creatives.
Studying the extent of adoption by advertisers and using this feedback to control creative generation is a promising direction for future research.

\bibliographystyle{ACM-Reference-Format}
\bibliography{refs}


\begin{thebibliography}{27}


\ifx \showCODEN    \undefined \def \showCODEN     #1{\unskip}     \fi
\ifx \showDOI      \undefined \def \showDOI       #1{#1}\fi
\ifx \showISBNx    \undefined \def \showISBNx     #1{\unskip}     \fi
\ifx \showISBNxiii \undefined \def \showISBNxiii  #1{\unskip}     \fi
\ifx \showISSN     \undefined \def \showISSN      #1{\unskip}     \fi
\ifx \showLCCN     \undefined \def \showLCCN      #1{\unskip}     \fi
\ifx \shownote     \undefined \def \shownote      #1{#1}          \fi
\ifx \showarticletitle \undefined \def \showarticletitle #1{#1}   \fi
\ifx \showURL      \undefined \def \showURL       {\relax}        \fi
\providecommand\bibfield[2]{#2}
\providecommand\bibinfo[2]{#2}
\providecommand\natexlab[1]{#1}
\providecommand\showeprint[2][]{arXiv:#2}

\bibitem[\protect\citeauthoryear{??}{mat}{[n.d.]}]%
        {matchzoo}
 \bibinfo{year}{[n.d.]}\natexlab{}.
\newblock \bibinfo{title}{Match Zoo}.
\newblock
  \bibinfo{howpublished}{\url{https://github.com/NTMC-Community/MatchZoo}}.
\newblock


\bibitem[\protect\citeauthoryear{??}{shu}{[n.d.]}]%
        {shutterstock}
 \bibinfo{year}{[n.d.]}\natexlab{}.
\newblock \bibinfo{title}{Shutterstock: Search millions of royalty free stock
  images, photos, videos, and music.}
\newblock \bibinfo{howpublished}{\url{https://www.shutterstock.com/}}.
\newblock


\bibitem[\protect\citeauthoryear{??}{tab}{[n.d.]}]%
        {taboola_trends}
 \bibinfo{year}{[n.d.]}\natexlab{}.
\newblock \bibinfo{title}{Taboola-trends}.
\newblock \bibinfo{howpublished}{\url{https://trends.taboola.com/}}.
\newblock


\bibitem[\protect\citeauthoryear{Bahdanau, Cho, and Bengio}{Bahdanau
  et~al\mbox{.}}{[n.d.]}]%
        {bahdanau2014neural}
\bibfield{author}{\bibinfo{person}{Dzmitry Bahdanau},
  \bibinfo{person}{Kyunghyun Cho}, {and} \bibinfo{person}{Yoshua Bengio}.}
  \bibinfo{year}{[n.d.]}\natexlab{}.
\newblock \showarticletitle{Neural Machine Translation by Jointly Learning to
  Align and Translate}. In \bibinfo{booktitle}{\emph{ICLR 2015}}.
\newblock


\bibitem[\protect\citeauthoryear{Bhamidipati, Kant, and Mishra}{Bhamidipati
  et~al\mbox{.}}{2017}]%
        {mappi_CIKM}
\bibfield{author}{\bibinfo{person}{Narayan Bhamidipati}, \bibinfo{person}{Ravi
  Kant}, {and} \bibinfo{person}{Shaunak Mishra}.}
  \bibinfo{year}{2017}\natexlab{}.
\newblock \showarticletitle{A Large Scale Prediction Engine for App Install
  Clicks and Conversions}. In \bibinfo{booktitle}{\emph{Proceedings of the 2017
  ACM on Conference on Information and Knowledge Management}}
  \emph{(\bibinfo{series}{CIKM '17})}. \bibinfo{pages}{167–175}.
\newblock


\bibitem[\protect\citeauthoryear{Boudin}{Boudin}{2016}]%
        {pke}
\bibfield{author}{\bibinfo{person}{Florian Boudin}.}
  \bibinfo{year}{2016}\natexlab{}.
\newblock \showarticletitle{pke: an open source python-based keyphrase
  extraction toolkit}. In \bibinfo{booktitle}{\emph{COLING 2016}}.
\newblock


\bibitem[\protect\citeauthoryear{Boudin}{Boudin}{2018}]%
        {multipartite_rank}
\bibfield{author}{\bibinfo{person}{Florian Boudin}.}
  \bibinfo{year}{2018}\natexlab{}.
\newblock \showarticletitle{Unsupervised Keyphrase Extraction with Multipartite
  Graphs}. In \bibinfo{booktitle}{\emph{ACL 2018}}.
\newblock


\bibitem[\protect\citeauthoryear{Florescu and Caragea}{Florescu and
  Caragea}{2017}]%
        {position-rank}
\bibfield{author}{\bibinfo{person}{Corina Florescu} {and}
  \bibinfo{person}{Cornelia Caragea}.} \bibinfo{year}{2017}\natexlab{}.
\newblock \showarticletitle{{P}osition{R}ank: An Unsupervised Approach to
  Keyphrase Extraction from Scholarly Documents}. In
  \bibinfo{booktitle}{\emph{ACL 2017}}.
\newblock


\bibitem[\protect\citeauthoryear{Guo, Fan, Ai, and Croft}{Guo
  et~al\mbox{.}}{2016}]%
        {drmm}
\bibfield{author}{\bibinfo{person}{Jiafeng Guo}, \bibinfo{person}{Yixing Fan},
  \bibinfo{person}{Qingyao Ai}, {and} \bibinfo{person}{W.~Bruce Croft}.}
  \bibinfo{year}{2016}\natexlab{}.
\newblock \showarticletitle{A Deep Relevance Matching Model for Ad-hoc
  Retrieval}. In \bibinfo{booktitle}{\emph{CIKM 2016}}.
\newblock


\bibitem[\protect\citeauthoryear{He, Liao, Zhang, Nie, Hu, and Chua}{He
  et~al\mbox{.}}{2017}]%
        {neural_collaborative_filtering}
\bibfield{author}{\bibinfo{person}{Xiangnan He}, \bibinfo{person}{Lizi Liao},
  \bibinfo{person}{Hanwang Zhang}, \bibinfo{person}{Liqiang Nie},
  \bibinfo{person}{Xia Hu}, {and} \bibinfo{person}{Tat-Seng Chua}.}
  \bibinfo{year}{2017}\natexlab{}.
\newblock \showarticletitle{Neural Collaborative Filtering}. In
  \bibinfo{booktitle}{\emph{WWW}}.
\newblock


\bibitem[\protect\citeauthoryear{Hughes, Chang, and Zhang}{Hughes
  et~al\mbox{.}}{[n.d.]}]%
        {microsoft_ad_generation_kdd19}
\bibfield{author}{\bibinfo{person}{J.~Weston Hughes}, \bibinfo{person}{Keng-hao
  Chang}, {and} \bibinfo{person}{Ruofei Zhang}.}
  \bibinfo{year}{[n.d.]}\natexlab{}.
\newblock \showarticletitle{Generating Better Search Engine Text Advertisements
  with Deep Reinforcement Learning} \emph{(\bibinfo{series}{KDD ’19})}.
\newblock


\bibitem[\protect\citeauthoryear{Hussain, Zhang, Zhang, Ye, Thomas, Agha, Ong,
  and Kovashka}{Hussain et~al\mbox{.}}{2017}]%
        {cvpr_kovashka}
\bibfield{author}{\bibinfo{person}{Zaeem Hussain}, \bibinfo{person}{Mingda
  Zhang}, \bibinfo{person}{Xiaozhong Zhang}, \bibinfo{person}{Keren Ye},
  \bibinfo{person}{Christopher Thomas}, \bibinfo{person}{Zuha Agha},
  \bibinfo{person}{Nathan Ong}, {and} \bibinfo{person}{Adriana Kovashka}.}
  \bibinfo{year}{2017}\natexlab{}.
\newblock \showarticletitle{Automatic Understanding of Image and Video
  Advertisements}. In \bibinfo{booktitle}{\emph{{CVPR}}}.
\newblock


\bibitem[\protect\citeauthoryear{Klein, Kim, Deng, Senellart, and Rush}{Klein
  et~al\mbox{.}}{2017}]%
        {opennmt}
\bibfield{author}{\bibinfo{person}{Guillaume Klein}, \bibinfo{person}{Yoon
  Kim}, \bibinfo{person}{Yuntian Deng}, \bibinfo{person}{Jean Senellart}, {and}
  \bibinfo{person}{Alexander~M. Rush}.} \bibinfo{year}{2017}\natexlab{}.
\newblock \showarticletitle{Open{NMT}}. In \bibinfo{booktitle}{\emph{ACL
  2017}}.
\newblock


\bibitem[\protect\citeauthoryear{Koren}{Koren}{2008}]%
        {koren_MF}
\bibfield{author}{\bibinfo{person}{Yehuda Koren}.}
  \bibinfo{year}{2008}\natexlab{}.
\newblock \showarticletitle{Factorization Meets the Neighborhood: A
  Multifaceted Collaborative Filtering Model}. In
  \bibinfo{booktitle}{\emph{KDD}}.
\newblock


\bibitem[\protect\citeauthoryear{Krasin, Duerig, Alldrin, Ferrari,
  Abu-El-Haija, Kuznetsova, Rom, Uijlings, Popov, Veit, Belongie, Gomes, Gupta,
  Sun, Chechik, Cai, Feng, Narayanan, and Murphy}{Krasin et~al\mbox{.}}{2017}]%
        {openimages}
\bibfield{author}{\bibinfo{person}{Ivan Krasin}, \bibinfo{person}{Tom Duerig},
  \bibinfo{person}{Neil Alldrin}, \bibinfo{person}{Vittorio Ferrari},
  \bibinfo{person}{Sami Abu-El-Haija}, \bibinfo{person}{Alina Kuznetsova},
  \bibinfo{person}{Hassan Rom}, \bibinfo{person}{Jasper Uijlings},
  \bibinfo{person}{Stefan Popov}, \bibinfo{person}{Andreas Veit},
  \bibinfo{person}{Serge Belongie}, \bibinfo{person}{Victor Gomes},
  \bibinfo{person}{Abhinav Gupta}, \bibinfo{person}{Chen Sun},
  \bibinfo{person}{Gal Chechik}, \bibinfo{person}{David Cai},
  \bibinfo{person}{Zheyun Feng}, \bibinfo{person}{Dhyanesh Narayanan}, {and}
  \bibinfo{person}{Kevin Murphy}.} \bibinfo{year}{2017}\natexlab{}.
\newblock \showarticletitle{OpenImages: A public dataset for large-scale
  multi-label and multi-class image classification.}
\newblock \bibinfo{journal}{\emph{Dataset available from
  https://github.com/openimages}} (\bibinfo{year}{2017}).
\newblock


\bibitem[\protect\citeauthoryear{Li, Wang, Zhang, Cui, Mao, and Jin}{Li
  et~al\mbox{.}}{2010}]%
        {explore_exploit_li}
\bibfield{author}{\bibinfo{person}{Wei Li}, \bibinfo{person}{Xuerui Wang},
  \bibinfo{person}{Ruofei Zhang}, \bibinfo{person}{Ying Cui},
  \bibinfo{person}{Jianchang Mao}, {and} \bibinfo{person}{Rong Jin}.}
  \bibinfo{year}{2010}\natexlab{}.
\newblock \showarticletitle{Exploitation and Exploration in a Performance Based
  Contextual Advertising System}. In \bibinfo{booktitle}{\emph{KDD 2010}}.
\newblock


\bibitem[\protect\citeauthoryear{Lin}{Lin}{2004}]%
        {ROUGE}
\bibfield{author}{\bibinfo{person}{Chin-Yew Lin}.}
  \bibinfo{year}{2004}\natexlab{}.
\newblock \showarticletitle{{ROUGE}: A Package for Automatic Evaluation of
  Summaries}. In \bibinfo{booktitle}{\emph{Text Summarization Branches Out}}.
  \bibinfo{publisher}{ACL}.
\newblock


\bibitem[\protect\citeauthoryear{Luong, Pham, and Manning}{Luong
  et~al\mbox{.}}{2015}]%
        {luong2015attention}
\bibfield{author}{\bibinfo{person}{Thang Luong}, \bibinfo{person}{Hieu Pham},
  {and} \bibinfo{person}{Christopher~D. Manning}.}
  \bibinfo{year}{2015}\natexlab{}.
\newblock \showarticletitle{Effective Approaches to Attention-based Neural
  Machine Translation}. In \bibinfo{booktitle}{\emph{EMNLP 2015}}.
\newblock


\bibitem[\protect\citeauthoryear{McMahan, Holt, Sculley, Young, Ebner, Grady,
  Nie, Phillips, Davydov, Golovin, Chikkerur, Liu, Wattenberg, Hrafnkelsson,
  Boulos, and Kubica}{McMahan et~al\mbox{.}}{[n.d.]}]%
        {Google_FTRL}
\bibfield{author}{\bibinfo{person}{H.~Brendan McMahan}, \bibinfo{person}{Gary
  Holt}, \bibinfo{person}{D. Sculley}, \bibinfo{person}{Michael Young},
  \bibinfo{person}{Dietmar Ebner}, \bibinfo{person}{Julian Grady},
  \bibinfo{person}{Lan Nie}, \bibinfo{person}{Todd Phillips},
  \bibinfo{person}{Eugene Davydov}, \bibinfo{person}{Daniel Golovin},
  \bibinfo{person}{Sharat Chikkerur}, \bibinfo{person}{Dan Liu},
  \bibinfo{person}{Martin Wattenberg}, \bibinfo{person}{Arnar~Mar
  Hrafnkelsson}, \bibinfo{person}{Tom Boulos}, {and} \bibinfo{person}{Jeremy
  Kubica}.} \bibinfo{year}{[n.d.]}\natexlab{}.
\newblock \showarticletitle{Ad Click Prediction: a View from the Trenches}
  \emph{(\bibinfo{series}{KDD 2013})}.
\newblock


\bibitem[\protect\citeauthoryear{Mishra, Verma, and Gligorijevic}{Mishra
  et~al\mbox{.}}{2019}]%
        {self_recsys2019}
\bibfield{author}{\bibinfo{person}{Shaunak Mishra}, \bibinfo{person}{Manisha
  Verma}, {and} \bibinfo{person}{Jelena Gligorijevic}.}
  \bibinfo{year}{2019}\natexlab{}.
\newblock \showarticletitle{Guiding Creative Design in Online Advertising}
  \emph{(\bibinfo{series}{RecSys '19})}. \bibinfo{pages}{418–422}.
\newblock


\bibitem[\protect\citeauthoryear{Papineni, Roukos, Ward, and Zhu}{Papineni
  et~al\mbox{.}}{2002}]%
        {BLEU}
\bibfield{author}{\bibinfo{person}{Kishore Papineni}, \bibinfo{person}{Salim
  Roukos}, \bibinfo{person}{Todd Ward}, {and} \bibinfo{person}{Wei-Jing Zhu}.}
  \bibinfo{year}{2002}\natexlab{}.
\newblock \showarticletitle{BLEU: A Method for Automatic Evaluation of Machine
  Translation}. \bibinfo{publisher}{ACL}.
\newblock


\bibitem[\protect\citeauthoryear{Pennington, Socher, and Manning}{Pennington
  et~al\mbox{.}}{2014}]%
        {glove}
\bibfield{author}{\bibinfo{person}{Jeffrey Pennington},
  \bibinfo{person}{Richard Socher}, {and} \bibinfo{person}{Christopher~D.
  Manning}.} \bibinfo{year}{2014}\natexlab{}.
\newblock \showarticletitle{Glove: Global vectors for word representation}. In
  \bibinfo{booktitle}{\emph{In EMNLP 2014}}.
\newblock


\bibitem[\protect\citeauthoryear{Schmidt and Eisend}{Schmidt and
  Eisend}{2015}]%
        {ad_fatigue_schmidt}
\bibfield{author}{\bibinfo{person}{Susanne Schmidt} {and}
  \bibinfo{person}{Martin Eisend}.} \bibinfo{year}{2015}\natexlab{}.
\newblock \showarticletitle{Advertising Repetition: A Meta-Analysis on
  Effective Frequency in Advertising}.
\newblock \bibinfo{journal}{\emph{Journal of Advertising}}
  (\bibinfo{year}{2015}).
\newblock


\bibitem[\protect\citeauthoryear{See, Liu, and Manning}{See
  et~al\mbox{.}}{[n.d.]}]%
        {see_pointer_generator}
\bibfield{author}{\bibinfo{person}{Abigail See}, \bibinfo{person}{Peter~J.
  Liu}, {and} \bibinfo{person}{Christopher~D. Manning}.}
  \bibinfo{year}{[n.d.]}\natexlab{}.
\newblock \showarticletitle{Get To The Point: Summarization with
  Pointer-Generator Networks}. In \bibinfo{booktitle}{\emph{ACL 2017}}.
\newblock


\bibitem[\protect\citeauthoryear{Yang, Lan, Guo, Fan, Zhu, Lan, Wang, and
  Cheng}{Yang et~al\mbox{.}}{[n.d.]}]%
        {drmm_topk}
\bibfield{author}{\bibinfo{person}{Zhou Yang}, \bibinfo{person}{Qingfeng Lan},
  \bibinfo{person}{Jiafeng Guo}, \bibinfo{person}{Yixing Fan},
  \bibinfo{person}{Xiaofei Zhu}, \bibinfo{person}{Yanyan Lan},
  \bibinfo{person}{Yue Wang}, {and} \bibinfo{person}{Xueqi Cheng}.}
  \bibinfo{year}{[n.d.]}\natexlab{}.
\newblock \showarticletitle{A Deep Top-K Relevance Matching Model for Ad-hoc
  Retrieval}. In \bibinfo{booktitle}{\emph{Information Retrieval - 24th China
  Conference, {CCIR} 2018}}.
\newblock


\bibitem[\protect\citeauthoryear{Zhou, Mishra, Gligorijevic, Bhatia, and
  Bhamidipati}{Zhou et~al\mbox{.}}{2019}]%
        {gemx_kdd}
\bibfield{author}{\bibinfo{person}{Yichao Zhou}, \bibinfo{person}{Shaunak
  Mishra}, \bibinfo{person}{Jelena Gligorijevic}, \bibinfo{person}{Tarun
  Bhatia}, {and} \bibinfo{person}{Narayan Bhamidipati}.}
  \bibinfo{year}{2019}\natexlab{}.
\newblock \showarticletitle{Understanding Consumer Journey using Attention
  based Recurrent Neural Networks}.
\newblock \bibinfo{journal}{\emph{KDD}} (\bibinfo{year}{2019}).
\newblock


\bibitem[\protect\citeauthoryear{Zhou, Mishra, Verma, Bhamidipati, and
  Wang}{Zhou et~al\mbox{.}}{2020}]%
        {www20_joey}
\bibfield{author}{\bibinfo{person}{Yichao Zhou}, \bibinfo{person}{Shaunak
  Mishra}, \bibinfo{person}{Manisha Verma}, \bibinfo{person}{Narayan
  Bhamidipati}, {and} \bibinfo{person}{Wei Wang}.}
  \bibinfo{year}{2020}\natexlab{}.
\newblock \showarticletitle{Recommending Themes for Ad Creative Design via
  Visual-Linguistic Representations}. In \bibinfo{booktitle}{\emph{Proceedings
  of The Web Conference 2020}} \emph{(\bibinfo{series}{WWW '20})}.
  \bibinfo{pages}{2521–2527}.
\newblock


\end{thebibliography}
\end{document}